\documentclass[10pt, twocolumn]{article}
\usepackage{ijcai18}

\usepackage{times}
\usepackage{xcolor}
\usepackage{soul}
\usepackage[utf8]{inputenc}
\usepackage[small]{caption}

\usepackage{latexsym} 

\usepackage{amsmath}
\usepackage{amsfonts}
\usepackage{booktabs}
\usepackage[font=small]{caption}
\usepackage{enumitem}
\usepackage[hang]{footmisc}
\usepackage{graphicx}
\usepackage[hidelinks]{hyperref} 
\usepackage{makecell}
\usepackage{multirow}
\usepackage{subcaption}
\usepackage{tabulary}
\usepackage[noindentafter]{titlesec}

\title{Virtual-to-Real: Learning to Control  in Visual Semantic Segmentation}

\author{
  \parbox{\linewidth}{\centering
    Zhang-Wei Hong,
    Yu-Ming Chen$^*$,
    Hsuan-Kung Yang$^*$,
    Shih-Yang Su$^*$,
    Tzu-Yun Shann$^*$,
    Yi-Hsiang Chang,
    Brian Hsi-Lin Ho,
    Chih-Chieh Tu,
    Tsu-Ching Hsiao,
    Hsin-Wei Hsiao,
    Sih-Pin Lai, 
    Yueh-Chuan Chang,
    Chun-Yi Lee%
  }%
\\
Elsa Lab, Department of Computer Science, National Tsing Hua University, Hsinchu, Taiwan
}

\begin{document}

\maketitle

\begin{abstract}
Collecting training data from the physical world is usually time-consuming and even dangerous for fragile robots, and thus, recent advances in robot learning advocate the use of simulators as the training platform. Unfortunately, the reality gap between synthetic and real visual data prohibits direct migration of the models trained in virtual worlds to the real world. This paper proposes a modular architecture for tackling the virtual-to-real problem. The proposed architecture separates the learning model into a perception module and a control policy module, and uses semantic image segmentation as the meta representation for relating these two modules.  The perception module translates each perceived RGB image to semantic image segmentation.  The control policy module is implemented as a deep reinforcement learning agent, which performs actions based on the translated image segmentation. Our architecture is evaluated in an obstacle avoidance task and a target following task.  Experimental results show that our architecture significantly outperforms all of the baseline methods in both virtual and real environments, and demonstrates a faster learning curve than them.  We also present a detailed analysis for a variety of variant configurations, and validate the transferability of our modular architecture.
{\let\thefootnote\relax\footnote{{
* indicates equal contribution.
}}}
\end{abstract}

\section{Introduction}
\label{sec::intro}
Visual perception based control has been attracting attention in recent years for controlling robotic systems, as visual inputs contain rich information of the unstructured physical world.  It is usually necessary for an autonomous robot to understand visual scene semantics to navigate to a specified destination. Interpreting and representing visual inputs to perform actions and interact with objects, however, are challenging for robots in unstructured environments as colored images are typically complex and noisy~\cite{maier2012real,biswas2012depth}. It is especially difficult to design a rule-based robot satisfying such requirements.

Both modular and end-to-end learning-based approaches have been proven effective in a variety of vision-based robotic control tasks~\cite{sadeghi2016cad,finn2017deep,gupta2017cognitive,smolyanskiy2017toward,zhu2017target}.  A modular cognitive mapping and planning approach has been demonstrated successful in first-person visual navigation~\cite{gupta2017cognitive}.  Vision-based reinforcement learning (RL) has been attempted to train an end-to-end control policy for searching specific targets~\cite{zhu2017target}.  Applying end-to-end supervised learning to navigate a drone along a trail with human-labeled image-action pairs is presented in~\cite{smolyanskiy2017toward}.  An end-to-end training methodology for object manipulation tasks using unlabeled video data is described in~\cite{finn2017deep}.  While these learning-based approaches seem attractive, they typically require a huge amount of training data.  Collecting training data for learning a control policy in the physical world is usually costly and poses a number of challenges.  First, preparing large amounts of labeled data for supervised learning takes considerable time and human efforts.  Second, RL relies on trial-and-error experiences, which restrict fragile robots from dangerous tasks. Online training and fine-tuning robots in the physical world also tend to be time-consuming, limiting the learning efficiency of various RL algorithms.

An alternative approach to accelerate the learning efficiency and reduce the cost is training robots in virtual worlds.  Most of the recent works on robot learning collect training data from simulators~\cite{james20163d,rusu2016sim,sadeghi2016cad,peng2017sim,tobin2017domain,zhu2017target}.  However, the discrepancies between virtual and real worlds prohibit an agent trained in a virtual world from being transferred to the physical world directly~\cite{james20163d}. 
Images rendered by low-fidelity simulators are unlikely to contain as much rich information as real ones.  Therefore, bridging the reality gap~\cite{tobin2017domain} has been a challenging problem both in computer vision and robotics.
Many research efforts have been devoted to tackling this problem by either domain adaption (DA)~\cite{rusu2016sim,ghadirzadeh2017deep,zhang2017sim} or domain randomization (DR)~\cite{sadeghi2016cad,peng2017sim,tobin2017domain,zhang2017sim}.  Both of these methods train agents by simulators.  DA fine-tunes simulator-trained models with real-world data.  DR, on the other hand, trains agents with randomized object textures, colors, lighting conditions, or even physical dynamics in virtual worlds.  The diversified simulated data enable the agents to generalize their models to the real world.  Unfortunately, collecting the real-world data required by DA for control policy learning is 
exceptionally time-consuming.  Although DR does not require real-world data, the technique lacks a systematic way to determine which parameters are required to be randomized.  Furthermore, DR requires considerably large number of training samples in the training phase to achieve acceptable performance.

Instead of training a vision-based control policy in an end-to-end fashion, we propose a new modular architecture to address the reality gap in vision domain.
We focus on the problem of transferring deep neural network (DNN) models trained in simulated virtual worlds to the real world for vision-based robotic control.  We propose to separate the learning model into a perception module and a control policy module, and use semantic image segmentation as the meta-state for relating these two modules.  These two modules are trained separately and independently.  Each of them assumes no prior knowledge of the other module.  The perception module translates RGB images to semantic image segmentations.  This module can be any semantic image segmentation models pre-trained on either commonly available datasets~\cite{cordts2016cityscapes,zhou2017scene} or synthetic images generated by simulators~\cite{richter2017playing}.
As annotated datasets for semantic segmentation are widely available nowadays and cross dataset domain adaptation has been addressed \cite{chen2017no}, training and fine-tuning a perception module for segmenting outdoor~\cite{cordts2016cityscapes} or indoor~\cite{zhou2017scene} scenarios 
have become straightforward.  The control policy module employs deep RL methods and takes semantic image segmentation as its input.  In the training phase, the control policy module only receives the image segmentation rendered by a low-fidelity simulators.  The RL agent interacts with the simulated environments and collects training data for the control policy module.  While in the execution phase, the control policy module receives image segmentation from the perception module, enabling the RL agent to interact with the real world.  As the image segmentations rendered by the simulator and those generated by the perception module are invariant~\cite{you2017virtual}, the control policy learned from the simulator can be transferred to the real world directly.  The proposed architecture provides better efficiency than the conventional learning-based methods, as simulators are able to render image segmentations at a higher frame rate, enabling the RL agent to collect training or trial-and-error experiences faster. It also allows the RL agent to learn faster, as semantic image segmentation contains less noise than raw images. We believe that our methodology is more general and applicable to different scenarios and tasks than the method proposed in~\cite{Yan2017virtualtoreal}, which uses a binary segmentation mask to bridge the visual and control modules. Using binary segmentation can result in a loss of semantic information, which are essential in a number of complex tasks. On the other hand, our proposed methodology not only preserves semantic information, but is also generalizable to complex scenarios.

Another advantage of the proposed architectures is its modularity.  Both the perception and control policy modules can be flexibly replaced in a plug-and-play fashion, as long as their meta-state representation (i.e., image segmentation) formats are aligned with each other.  
Replacing the perception module allows a pre-trained control policy module to be applied to different scenarios.  Replacing the control policy module, on the other hand, changes the policy of the robot in the same environment.  
Furthermore, our architecture allows another visual-guidance module to be incorporated.  The visual-guidance module adjusts the meta-state representation, such that the behavior of the robot can be altered online without replacing either the perception or control policy modules.  
   
To demonstrate the effectiveness of the proposed modular architecture, we evaluate it against a number of baseline methods in two benchmark tasks: obstacle avoidance and target following. These tasks are essential for autonomous robots, as the former requires collision-free navigation, while the latter requires understanding of high-level semantics in the environments.
To validate the transferability of our architecture, we train our models and the baseline methods in our simulator, and compare their performance in the benchmark tasks in both virtual and real worlds. We focus on RL-based robotic control policy, though the control policy module can be implemented by various options (e.g., imitation learning). Our results show that the proposed architecture outperforms all the baseline methods in all the benchmark tasks. Moreover, it enables our RL agent to learn dramatically faster than all the baseline methods.  By simply replacing the perception module, we also show that no additional fine-tuning is necessary for the control policy module when migrating the operating environment.  We further demonstrate the versatility of the visual-guidance module in Section~\ref{sec::exp}. 
The contributions of this work are summarized as follows:
\begin{enumerate}
\item {A straightforward, easy to implement, and effective solution (which has not yet been proposed in the literature) to the challenging virtual-to-real transferring problem.}
\item A new modular learning-based architecture which separates the vision-based robotic learning model into a perception module and a control policy module.
\item A novel concept for bridging the reality gap via the use of semantic image segmentation, which serves as the meta-state representation for relating the two modules.
\item A way to migrate the operating environment of a robot to another one without further fine-tuning its control policy module.

\item A visual-guidance module for altering the behavior of a robot via adjusting the meta-state representation.
\end{enumerate}

The remainder of this paper is organized as follows.  Section~\ref{sec::background} introduces background material.  Section~\ref{sec::proposed_methodology} walks through the proposed modular architecture. Section~\ref{sec::evaluation-setup} describes the evaluation setup, tasks, and scenarios.  Section~\ref{sec::exp} presents the experimental results.  Section~\ref{sec::conclusion} concludes.

\section{Background}
\label{sec::background}
This section briefly reviews background material on semantic image segmentation, RL, and policy gradient methods.

\subsection{Semantic Image Segmentation}
\label{subsec::img_seg}
The goal of semantic segmentation is to perform dense predictions at pixel level. 
It has received great attention due to its applicability to a number of research domains (e.g., visual understanding, autonomous driving, and robotic control).
Fully convolutional network (FCN)~\cite{long2015fully} 
pioneered the replacement of fully-connected (FC) layers by convolutional layers.
A number of successive works have further enhanced the accuracy and efficiency~\cite{chen2016deeplab,zhao2017icnet,zhao2017pspnet,paszke2016enet,he2015spatial},
making them more promising for robotic control tasks. Unfortunately,
relatively fewer attempts have been made to incorporate them into robotic learning models.

\subsection{RL and Policy Gradient Methods}
\label{subsec::rl}
RL trains an agent to interact with an environment $\mathcal{E}$. An RL agent observes a state $s$ from $\mathcal{E}$, takes an action $a$ according to its policy $\pi(a|s)$, and receives a reward $r(s,a)$. $\mathcal{E}$ then transitions to a new state $s^\prime$. The agent's objective is to maximize its accumulated rewards $G_t$ with a discounted factor $\gamma$, expressed as $G_t=\sum^{T}_{\tau = t}\gamma^{\tau-t}r(s_{\tau},a_{\tau})$, where $t$ is the current timestep, and $T$ is the total number of timesteps. Policy gradient methods are RL approaches that directly optimize $\pi$ in the direction: 
$\small
\nabla_\pi \sum^T_{t=0} \log\pi(a_t|s_t) (G_t-b(s_t))$,
where $b(s_t)$ is a baseline function.
A common choice for $b(s_t)$ is the 
value function $V^{\pi}(s) = \mathbb{E}\big[G_{t}|s_{t} = s, \pi]$. This approach is known as the actor-critic algorithm~\cite{williams1992simple}. An asynchronous variant of the actor-critic algorithm, namely 
asynchronous advantage actor-critic (A3C)~\cite{mnih2016asynchronous}, has been proven data-efficient and suitable for vision-based control tasks~\cite{wu2016training,parisotto2017neural,you2017virtual,zhu2017target}.

\section{Proposed Methodology}
\label{sec::proposed_methodology}
\begin{figure}[t]
\centering
\includegraphics[width=1.0\linewidth]{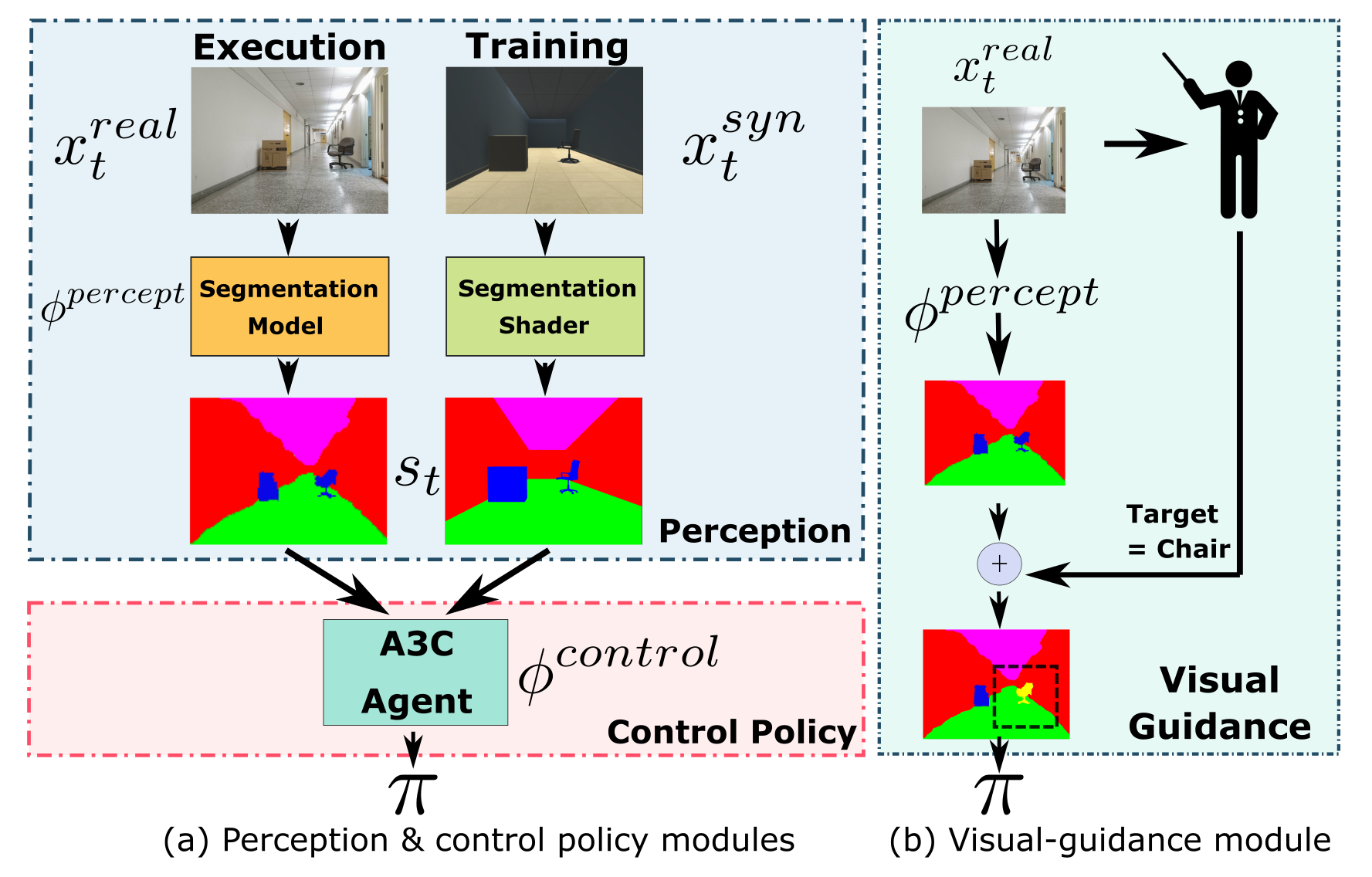}
\caption{Model architecture}
\label{figure::archi_overall}
\end{figure}

In this section, we present the modular architecture of our model.
We first provide an overview of the model architecture, followed by the implementation details of the 
modules. Fig.~\ref{figure::archi_overall} illustrates the proposed modular architecture. It consists of a perception module, a control policy module, and a visual guidance module. Semantic image segmentation $s_t$ serves as the meta-state representation for relating the former two modules, as shown in Fig.~\ref{figure::archi_overall}~(a). The perception module generates $s_t$ from an RGB input image $x_t$, which comes from different sources in the training ($x^{syn}_t$) and execution ($x^{real}_t$) phases. The control policy module takes $s_t$ as its input, and reacts with $a_t$ according to $\pi$. The visual guidance module enables high-level planning by modifying $s_t$,
as shown in Fig.~\ref{figure::archi_overall}~(b).


\subsection{Perception Module}
\label{subsec::percept}
The main function of the perception module is to generate $s_t$, and
passes it to the control policy module. In the training stage, $s_t$ is rendered by a segmentation shader from a synthetic image $x^{syn}_t$. While in the execution phase, the perception module can be any semantic segmentation model. Its function is expressed as $s_t = \phi^{percept}(x^{real}_t ; \theta^{percept})$, where $\phi^{percept}$ represents the semantic segmentation model, $x^{real}_t$ is the RGB image from the real world, and $\theta^{percept}$ denotes the parameters of the perception module. We directly employ existing models, such as DeepLab~\cite{chen2016deeplab} and ICNet~\cite{zhao2017icnet}, and pre-train them on datasets ADE20K~\cite{zhou2017scene} and Cityscape~\cite{cordts2016cityscapes} for indoor and outdoor scenarios, respectively. The format of meta-state representation $s_t$ is the same for both phases. 
The effect of using different semantic image segmentation models as the perception module is analyzed in Section~\ref{sec::exp}.

The benefit of using semantic segmentation as the meta-state representation
is threefold. First, although the visual appearances of $x^{real}_t$ and $x^{syn}_t$
differ, their semantic segmentations are almost identical~\cite{you2017virtual} (Fig.~\ref{figure::archi_overall}~(a)). This allows us to use semantic segmentation to bridge the reality gap.
Second, semantic segmentation preserves only the most important information of objects in an RGB image, resulting in a more structured representation. It also provides much richer information than other sensory modalities, as
they typically fail to cover crucial high-level scene semantics. 
Third, 
the perception module in the execution phase only requires a monocular camera,
which is a lot cheaper than depth cameras or laser rangefinders. 
As a result, semantic segmentation is inherently an ideal option for meta-state representation.

\subsection{Control Policy Module}
\label{subsec::control}
The control policy module is responsible for finding the best control policy $\pi$ for a specific task. In this work, it employs the A3C algorithm, and takes $s_t$ as its inputs (Fig.~\ref{figure::archi_overall}~(a)). In the training phase, the model $\phi^{control}$ is trained completely in simulated environments, and learns $\pi$ directly from $s_t$ rendered by a low-fidelity simulator. While in the execution phase, it receives $s_t$ from the perception module, allowing the RL agent to interact with the real world. As semantic segmentation is used as the meta-state representation,
it enables virtual-to-real transferring of $\pi$ without fine-tuning or DR.

\subsection{Visual Guidance Module}
\label{subsec::flexibility}
The use of semantic segmentation as the meta-state representation gives the proposed architecture extra flexibility.  Our modular architecture allows a visual-guidance module to be augmented to guide the control policy module to perform even more complex tasks by manipulating the meta-state $s_t$.  The manipulation can be carried out by either human beings or other RL agents.  One can easily modify a robot's task 
through manipulating the class labels in $s_t$. This implies that a roadway navigation robot can be transformed to a sidewalk navigation robot by swapping the labels of the roadway and the sidewalk in $s_t$. The visual guidance module can also alter a target following robot's objective online by modifying the target label to a new one, as shown in~Fig.~\ref{figure::archi_overall}~(b) (yellow chair). Note that visual guidance does not require any retraining, fine-tuning, or extra data in the above scenarios, while the previous works~\cite{gupta2017cognitive,zhu2017target} demand additional training samples to achieve the same level of flexibility.

\section{Virtual-to-Real Evaluation Setup}
\label{sec::evaluation-setup}

In this section, we present the evaluation setup in detail.  We evaluate the proposed  architecture both in virtual and real environments. We begin with
explaining the simulator used for training and evaluation, followed by an overview of the tasks and scenarios for performance evaluation. Finally, we discuss the baseline models and their settings. 

\setcounter{footnote}{0}

\paragraph{Simulator.}
We use the Unity3D \footnote{Unity3D: https://unity3d.com/} engine to develop the virtual environments for training our RL agent. As Unity3D supports simulation of rigid body dynamics (e.g., collision, friction, etc.), it allows virtual objects to
behave in a realistic way. In a simulated environment, our RL agent receives its observations in the form of semantic segmentation from a first-person perspective. It then determines the control action to take accordingly. The virtual environment responds to the action by transitioning to the next state and giving a reward signal. The geometric shape of our simulated agent is set to be similar to that of the real robot.

\paragraph{Tasks.}
The proposed model is evaluated against the baselines in virtual and real environments on two benchmark tasks: obstacle avoidance and target following. In both tasks, an RL agent starts at a random position in the environment at the beginning of each episode. At each timestep, it selects an action from the three possible options: moving forward, turning left, and turning right. Each action corresponds to the linear and angular velocities of the robot specified in Table~\ref{table::action_mapping}.

\begin{table}[!t]
    \centering
    \renewcommand{\arraystretch}{1.25}
    \begin{tabulary}{1\linewidth}{C|C} \toprule
	Action	& (Linear (m/s), Angular (rad/s)) \\ \hline
	Forward					& $(0.2, 0.0)$ \\ \hline
	TurnLeft				& $(0.2, -0.2)$ \\ \hline
	TurnRight				& $(0.2, 0.2)$ \\ \bottomrule
	\end{tabulary}
    \caption{Mapping actions to robot velocities.}
    \label{table::action_mapping}
\end{table}

\begin{figure}[t]
\centering
\includegraphics[width=.9\linewidth]{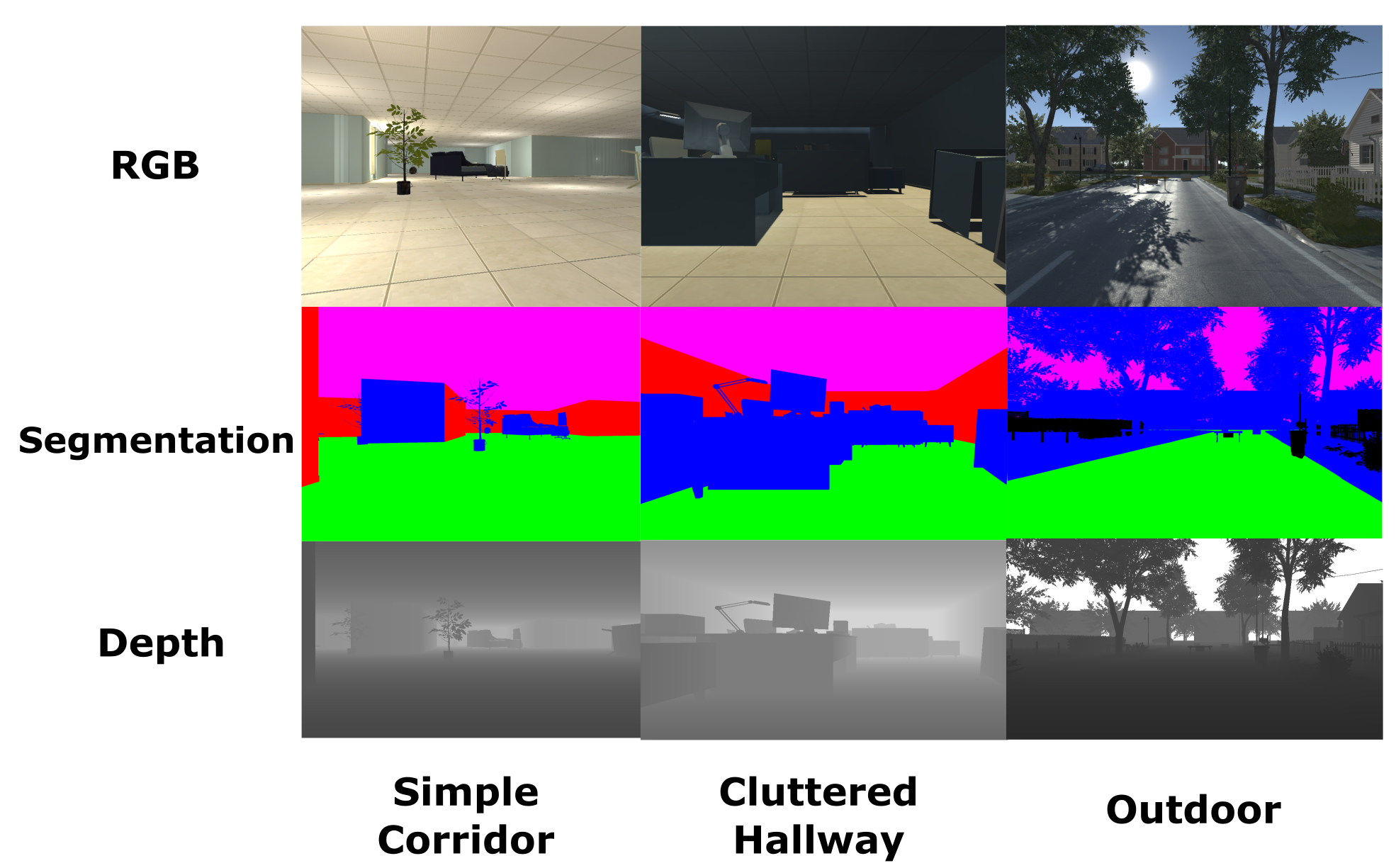}
\caption{Samples of evaluation scenarios. From left to right: \textit{Simple Corridor}, \textit{Cluttered Hallway}, and \textit{Outdoor}. From top to bottom: the corresponding RGB image, semantic segmentation, and depth map.}
\label{figure::sample_eval}
\end{figure}

\noindent \textit{(1) Obstacle Avoidance:}
The agent's goal is to navigate in a diverse set of scenarios, and avoid colliding with obstacles. 

\noindent \textit{(2) Target Following:} 
The agent's objective is to follow a moving target (e.g., human) while avoiding collisions. \\
We validate the proposed methodology on these simple tasks, as we consider our work as a proof-of-concept of transferring deep reinforcement learning (DRL) agents from virtual worlds to the real world by the use of semantic image segmentation. To the best of our knowledge and a rigorous literature survey, our work is the first one to address the virtual-to-real control problem with semantic image segmentation as the meta-state representation. Hence, we focused our attention on the effectiveness and efficiency of the proposed methodology, and validated it on the obstacle avoidance and target following tasks.

\paragraph{Scenarios.} We evaluate the models with the following three scenarios. Fig.~\ref{figure::sample_eval} illustrates a few sample scenes of them.\\
\noindent \textit{(1) Simple Corridor:} This scenario features indoor straight passages, sharp turns, static obstacles (e.g., chairs, boxes, tables, walls, etc.), and moving obstacles (e.g., human). 

\noindent \textit{(2) Cluttered Hallway:} This scenario features a hallway crammed with static and moving obstacles for evaluating an agent's capability of avoiding collisions in narrow space.

\noindent \textit{(3) Outdoor:} This scenario features an outdoor roadway with sidewalks, buildings, terrain, as well as moving cars and pedestrians. This is used to evaluate how well the control policy trained for the tasks can be transfered from an indoor environment to an outdoor environment. Note that the agent is not allowed to move on the sidewalks in this scenario.

\paragraph{Models.}
\label{subsubsec::model}

For all the experiments, our model is trained with semantic segmentations generated from the simulator. We adopt DR, depth perception, and ResNet~\cite{he2016res} as the baseline models. For the DR baseline, the texture of each mesh in a scene is randomly chosen from 100 textures. We augment the 100 textures by randomly changing their color tones and lighting conditions, which results in over 10,000 visually different textures. This ensures that the DR baseline used in our evaluation is actually trained with many different rendering conditions. For the depth perception baseline, we use the depth maps generated from the simulator. We set the minimum and maximum sensible depths of the agent in the simulator to be 0.3m and 25m, respectively, which are close to those of the depth cameras (e.g., ZED) mounted on real robots. We summarize the detailed settings of the model inputs in Table~\ref{table::model_setting}. All the models in Table~\ref{table::model_setting} adopt the vanilla A3C architecture~\cite{mnih2016asynchronous} except for ResNet-A3C, in which the convolutional layers are replaced by ResNet-101 pre-trained on ImageNet~\cite{krizhevsky2012imagenet}. ResNet-A3C is mainly used to justify that directly applying features extracted by ResNet as the meta-state representation does not lead to the same performance as ours.  In Table~\ref{table::model_setting}, we denote our model as Seg, and the other baselines as DR-A3C, Depth-A3C, and ResNet-A3C, respectively. The models named with "-S" (e.g., DR-A3C-S) indicate that these models  concatenate the latest four frames as their inputs.
We do not include DA in our experiments, as we only focus on control policies solely trained in simulated environments without further fine-tuning.
We use Adam optimizer~\cite{kingma2014adam}, and set both the learning rate and epsilon to 0.001. Each model is trained for 5M frames, and the training data are collected by 16 worker threads for all experimental settings. For the evaluation tasks in the real world, we train DeepLab-v2~\cite{chen2016deeplab} on ADE20K~\cite{zhou2017scene} and ICNet~\cite{zhao2017icnet} on Cityscapes~\cite{cordts2016cityscapes} as the indoor and outdoor perception modules, respectively. The class labels in ADE20K
are reduced as Table~\ref{table::class_label_mapping} for better accuracy and training efficiency.

\begin{table}
    \centering
    \renewcommand{\arraystretch}{1.25}
    \begin{tabulary}{1\linewidth}{C|C|C} \toprule
	Model			& Dimension									& Format \\ \hline
	Seg (Ours)		& $84$$\times$$84$$\times$$3$				& \multirowcell{2}{RGB\\Frame} \\
	Seg-S (Ours)	& $84$$\times$$84$$\times$$3$$\times$$4$	& \\ \hline
	DR-A3C			& $84$$\times$$84$$\times$$3$				& \multirowcell{2}{RGB\\Frame} \\
	DR-A3C-S		& $84$$\times$$84$$\times$$3$$\times$$4$	& \\ \hline
	Depth-A3C		& $84$$\times$$84$$\times$$1$				& \multirowcell{2}{Depth\\Map} \\
	Depth-A3C-S		& $84$$\times$$84$$\times$$4$				& \\ \hline
	ResNet-A3C		& $224$$\times$$224$$\times$$3$	& RGB Frame \\ \bottomrule
	\end{tabulary}
	\caption{Settings of the model inputs.}
	\label{table::model_setting}
\end{table}

\begin{table}[t]
\centering
\begin{minipage}[t]{0.9\linewidth}
	\renewcommand{\arraystretch}{1.25}
	\centering
	\scriptsize
	\begin{tabulary}{1\linewidth}{C|C} \toprule
	Reduced class labels		& \multirow{2}{*}{Original class labels in the ADE20K dataset} \\ \hline
	\multirow{2}{*}{Wall}		& Window, Door, Fence, Pillar, Signboard, Bulletin board \\ \hline
	Floor						& Road, Ground, Field, Path, Runway \\ \hline
	Ceil (Background)			& \multirow{2}{*}{Ceil} \\ \hline
	\multirow{2}{*}{Obstacle}	& Bed, Cabinet, Sofa, Table, Curtain, Chair, Shelf, Desk, Plant \\ \hline
	Target						& (Depends on scenario) \\ \bottomrule
	\end{tabulary}
	\caption{Class label reduction from the ADE20K dataset.}
	\label{table::class_label_mapping}
\end{minipage}
\end{table}

\begin{figure}[t]
\centering
\centering
\includegraphics[width=0.95\linewidth]{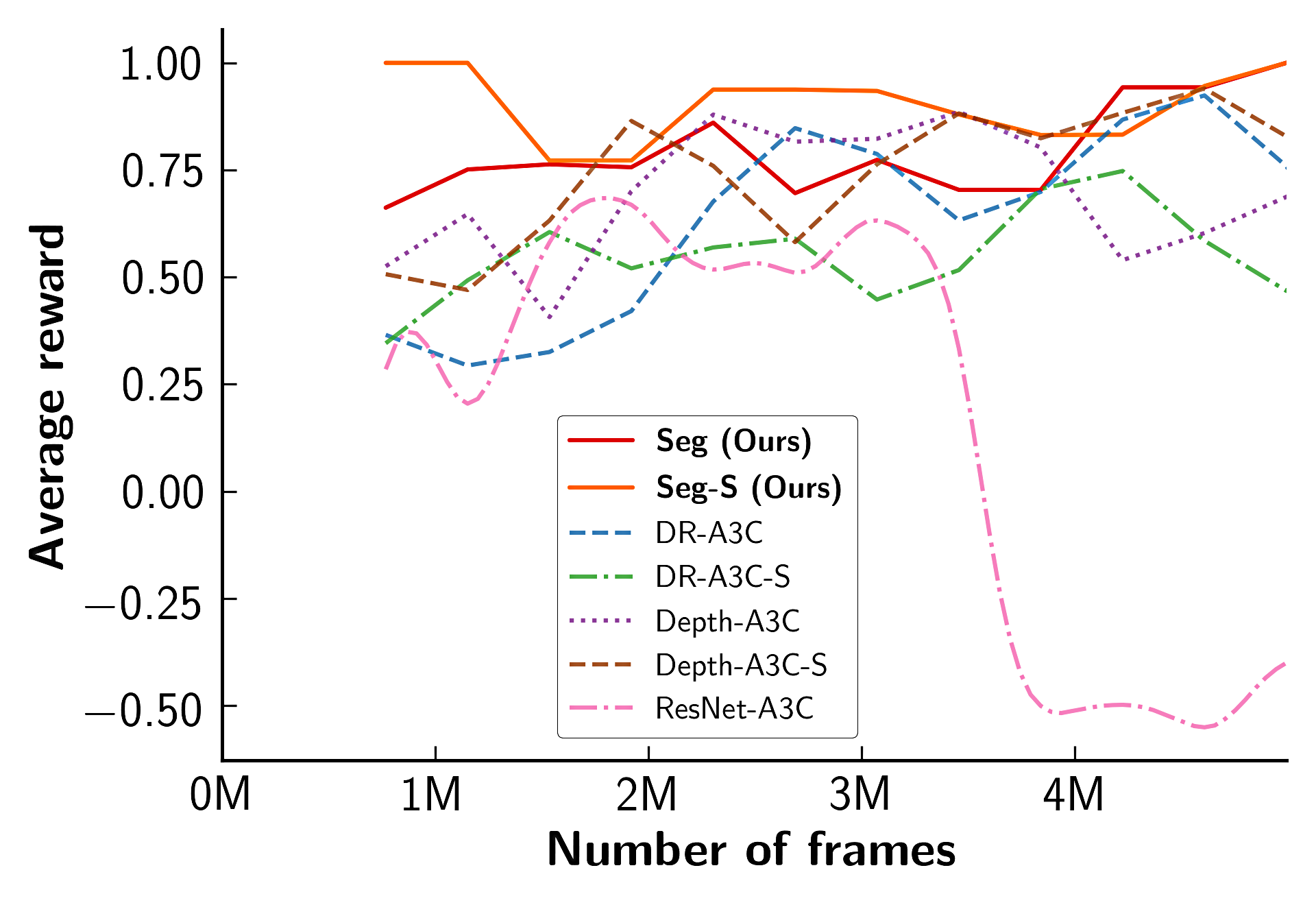}
\caption{Learning curves in the obstacle avoidance task.}
\label{figure::learning_curve_obstacle}
\end{figure}

\begin{figure}[t]
\centering
\includegraphics[width=0.9\linewidth]{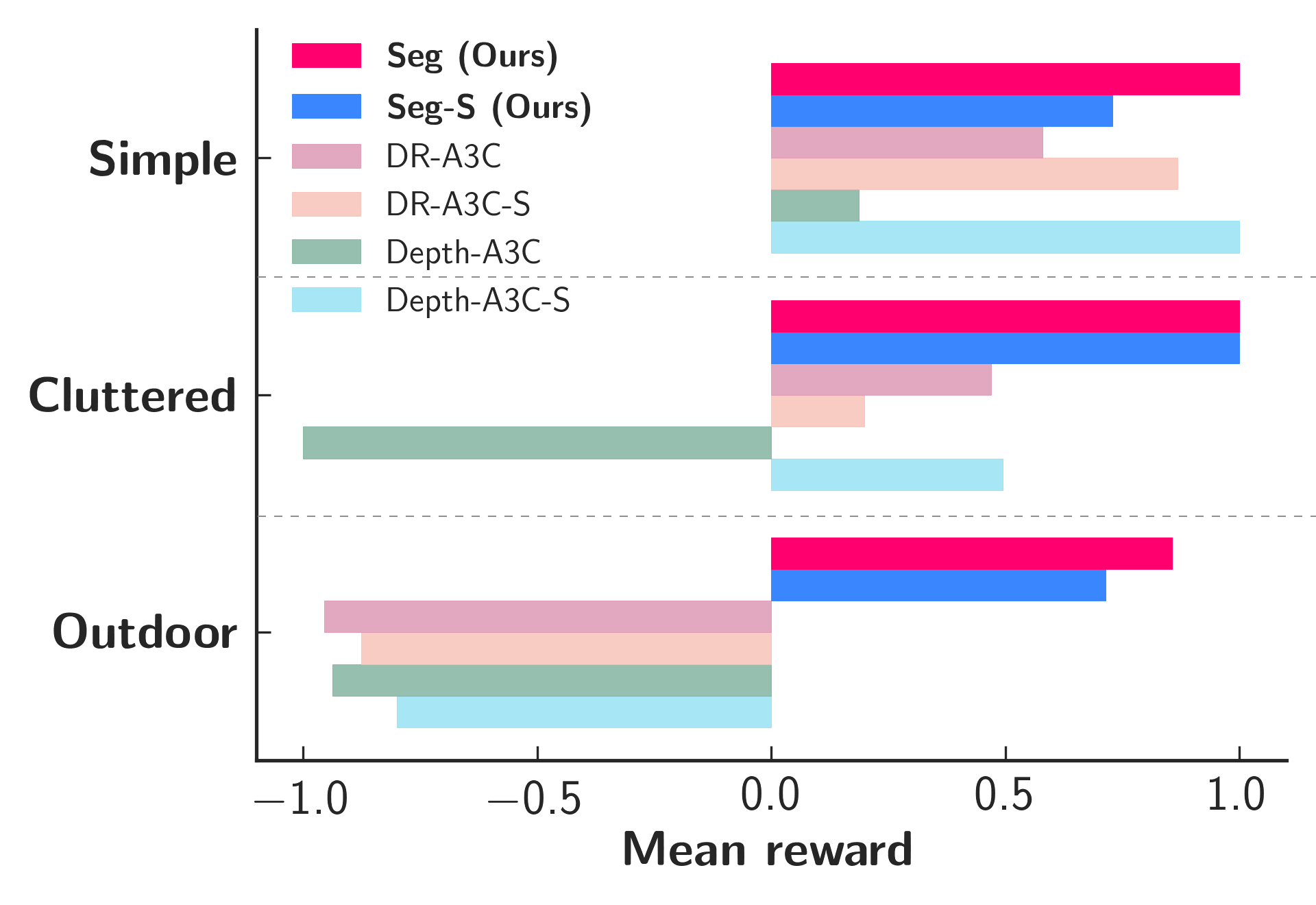}
\captionof{figure}{Evaluation results in the obstacle avoidance task (evaluated in simulated environments).}
\label{figure::mean_reward_obstacle_eval}
\end{figure}

\section{Experimental Results and Analysis}
\label{sec::exp}
In this section, we present the experimental results and their implications.  We comprehensively analyze the results, and perform an ablative study of our methodology. In Sections~\ref{subsec::obstacle_avoidance} and~\ref{subsec::target_following}, we compare the performance of our architecture and the baseline models in the tasks mentioned in Section~\ref{sec::evaluation-setup}. We demonstrate the concept of visual guidance in Section~\ref{subsec::visualguide}. We present the results of the ablative analysis in Section ~\ref{subsec::ablation}.

\subsection{Comparison in the Obstacle Avoidance Task}
\label{subsec::obstacle_avoidance}
To compare the performance of our models against those of the baselines in the obstacle avoidance tasks, we designed a total of 16 indoor scenarios for training the agents. Each scenario features a pre-defined set of attributes (e.g., hallway, static obstacles), along with randomly located obstacles and moving objects. A scenario is randomly selected for each training episode. Each episode terminates after 1,000 timesteps, or when a collision occurs. The agent receives a reward of 0.001 at each timesteps. Note that all the models are trained in simulation. Fig.~\ref{figure::learning_curve_obstacle} plots the learning curves of the models
in the obstacle avoidance task.
While most of the models achieve nearly optimal performance at the end of the training phase, our models learn significantly faster than the baseline models. This is due to the fact that semantic segmentation simplifies the original image representation to a more structured representation. On the contrary, the poor performance of ResNet-A3C indicates that the deep feature representations extracted by ResNet does not necessarily improve the performance. In addition, we observe that frame stacking has little impact on  performance in the training phase. Depth-A3C shows a slightly better performance than DR-A3C and ResNet-A3C, implying the usefulness of depth information. 

\paragraph{Evaluation in Simulation.} 

Fig.~\ref{figure::mean_reward_obstacle_eval} compares the mean rewards of the agents in three scenarios for seven types of models. The mean rewards are evaluated over 100 episodes, and the scenarios are totally different from those used at the training phase. It can be observed that our methods outperform the other baseline models in all of the three scenarios. In \textit{Simple Corridor}, noticeable performance can be observed from the other models. Most of the agents navigate well along the corridor, where there are relatively fewer obstacles compared to the other scenarios. In a more challenging scenario \textit{Cluttered Hallway}, it can be seen that both Depth-A3C and Depth-A3C-S experience a considerable drop in performance.
This is because depth map becomes too noisy in resolution to navigate in narrow space. In \textit{Outdoor}, all models receive negative mean rewards, except for our models. We observe that the baseline models tend to drive on the sidewalks, or rotate in place. In contrast, our models focus on driving on the roadway. As a result, we conclude that using semantic segmentation as the meta-state representation enables a control policy to be transfered from an indoor to an outdoor environment.

\begin{figure}[t]
\centering
\includegraphics[width=0.95\linewidth]{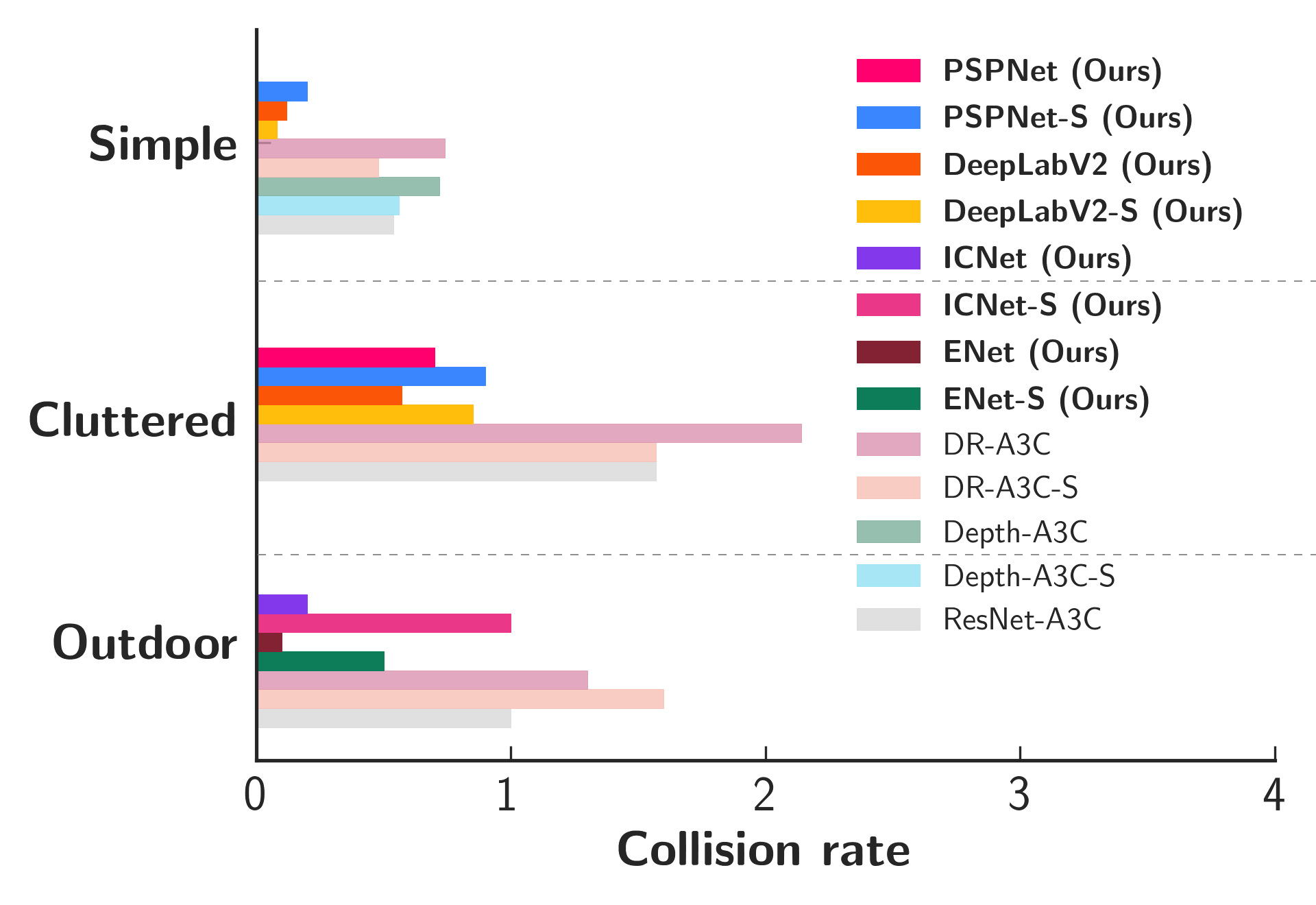}
\captionof{figure}{Collision rate in the obstacle avoidance task (evaluated in the real world).}
\label{figure::nav_collision}
\end{figure}

\begin{figure}[t]
\centering
\includegraphics[width=0.95\linewidth]{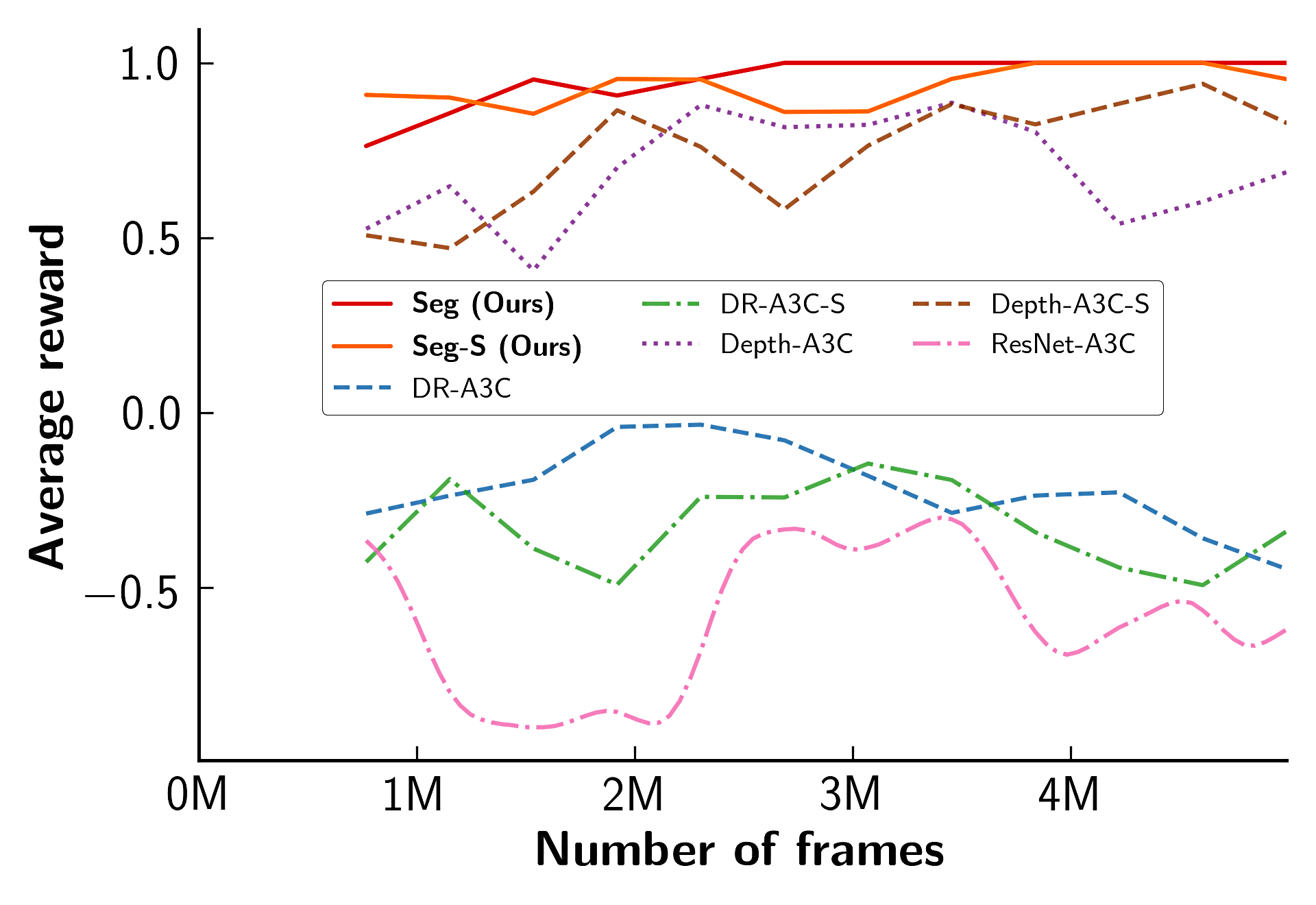}
\caption{Learning curves in the target following task}
\label{figure::learning_curve_target_following}
\end{figure}

\paragraph{Evaluation in the Real World.}
To test the virtual-to-real transferability of the models, we replicate the settings of the scenarios mentioned above in the real world. We measure the performance by recording the number of collisions per minute. We manually position the robot back to a random position when it collides with an obstacle. We report the results in Fig.~\ref{figure::nav_collision}. In \textit{Simple Corridor}, we observe that Depth-A3C experience a huge performance drop from simulation to the real world, since the depth map estimated in the real world is too noisy for the model to adapt. The performances of the other baselines are similarly decreased, while our methods still maintain the same level of performance. We exclude Depth-A3C from the rest of the real world scenarios due to the poor quality of depth estimation of the depth cameras in complex environments. In \textit{Cluttered Hallway}, while all of the models degrade in performance, our methods still significantly outperform the baselines. In \textit{Outdoor}, we observe that both DR-A3C and ResNet-A3C fail to transfer their knowledge from simulation to the real world, and tend to bump into cars and walls. We also notice that frame stacking leads to lower performance, because it amplifies the errors resulting from the changes in environmental dynamics. Additionally, to demonstrate the robustness of the proposed architecture to the quality of segmentation, we compare the performance of our methods with different segmentation models. They include PSPNet \cite{zhao2017pspnet} (high) and DeepLab \cite{chen2016deeplab} (low) for indoor environments, and ICNet \cite{zhao2017icnet} (high) and ENet \cite{paszke2016enet} (low) for outdoor environments. Fig.~\ref{figure::nav_collision} shows that our agents are not quite sensitive to the quality of image segmentations. Note that in \textit{Outdoor} scenario, ICNet performs worse than ENet due to its inevitable computational delay. From the results in Figs.~\ref{figure::learning_curve_obstacle}-\ref{figure::nav_collision}, we conclude that our methods are robust, generalizable, and transferable in the obstacle avoidance tasks.

\begin{figure}[t]
\centering
\includegraphics[width=0.85\linewidth]{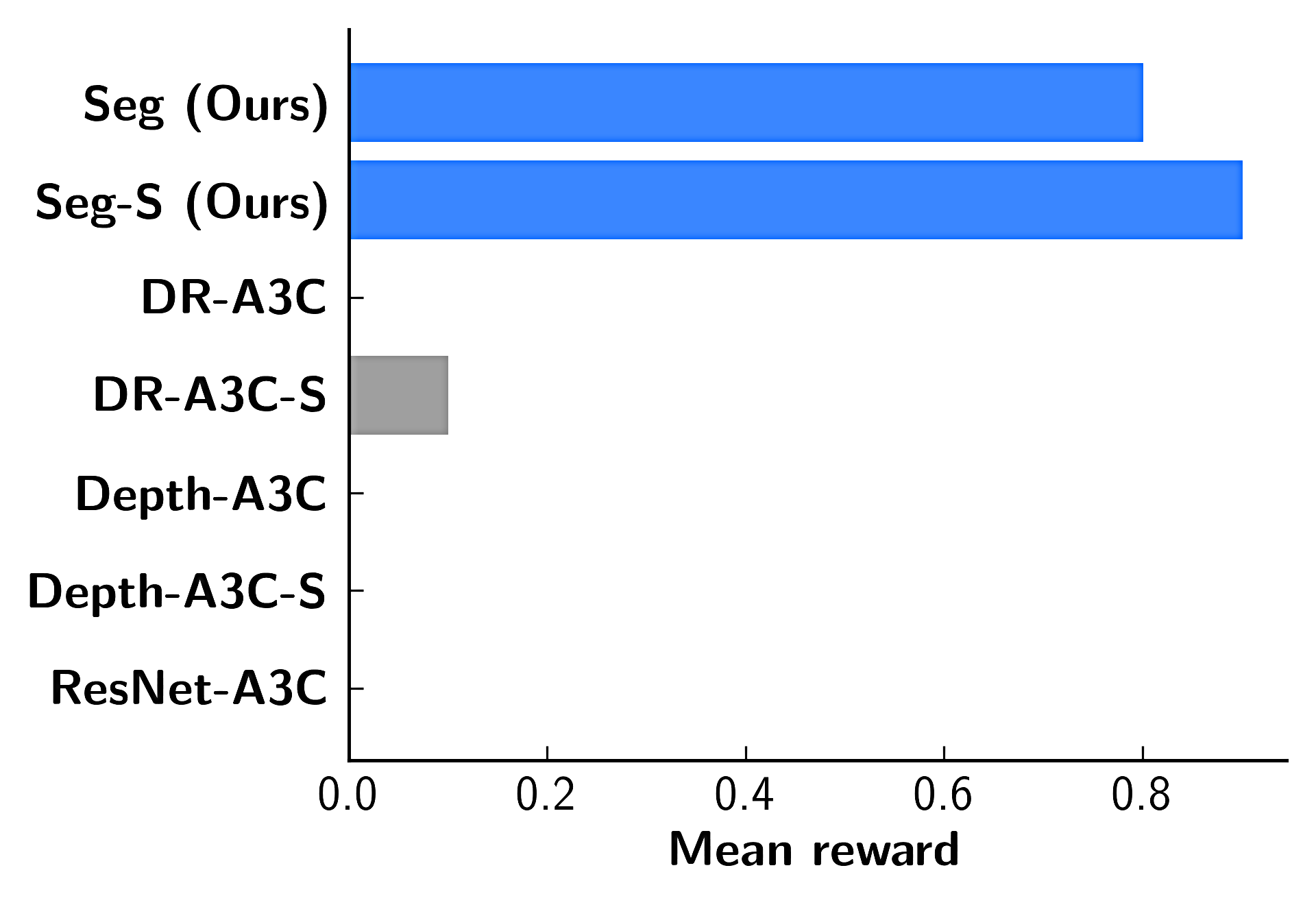}
\captionof{figure}{Evaluation results in the target following task (evaluated in simulated environments).}
\label{figure::mean_reward_follow_sim}
\end{figure}

\subsection{Comparison in the Target Following Task}
\label{subsec::target_following}
We further compare the performance of our methods against those of the baselines in the target following task. For each episode, a simulated scenario featuring static and moving obstacles is similarly selected at random. An episode ends immediately after 1,000 timesteps, or when a collision with the target/obstacles occurs. The agent receives a reward of 0.001 at each timestep if the target is within its sight, (1.0-``cumulative reward") if it touches the target, and 0.0 otherwise. At the beginning of each episode, the target is located in front of the agent. The target then navigates to a randomly chosen destination along the path computed by the A{*} algorithm. The ultimate goal of the agent is to avoid any collision on the way, and keep up with the target. For a fair comparison, the geometry of the target is always fixed. Fig.~\ref{figure::learning_curve_target_following} plots the learning curves of our models and the baselines in the training phase. It can be observed that our models are much superior to the other baseline models. We also notice that our agents learn to chase the target in the early stages of the training phase, while the baseline models never learn to follow the target. We conclude that the superiority of our models over the baselines results from our meta-state representation, which makes the target easily recognizable by our agents.


\paragraph{Evaluation in Simulation.}
To validate the generalizability, we evaluate the agent in a virtual scenario, which is not included in the training scenario sets. In each episode, the agent starts from a random position in the scenario.  We evaluate the mean rewards over 100 episodes, and present the results in Fig.~\ref{figure::mean_reward_follow_sim}. Our models achieve higher mean rewards than all the baselines, indicating the effectiveness of our models in capturing and tracking a fixed moving object. We observe that DR-A3C and ResNet-A3C always fail to follow the target at the fork of a corridor, because they are easily distracted by other objects. On the other hand, although Depth-A3C has higher mean rewards than the other baselines in the training phase, it often fails when the target is far away from it in the evaluation phase. The results in Fig.~\ref{figure::mean_reward_follow_sim} validate that our models are generalizable and transferable to new scenarios.

\begin{figure}[t]
\centering
\includegraphics[width=0.85\linewidth]{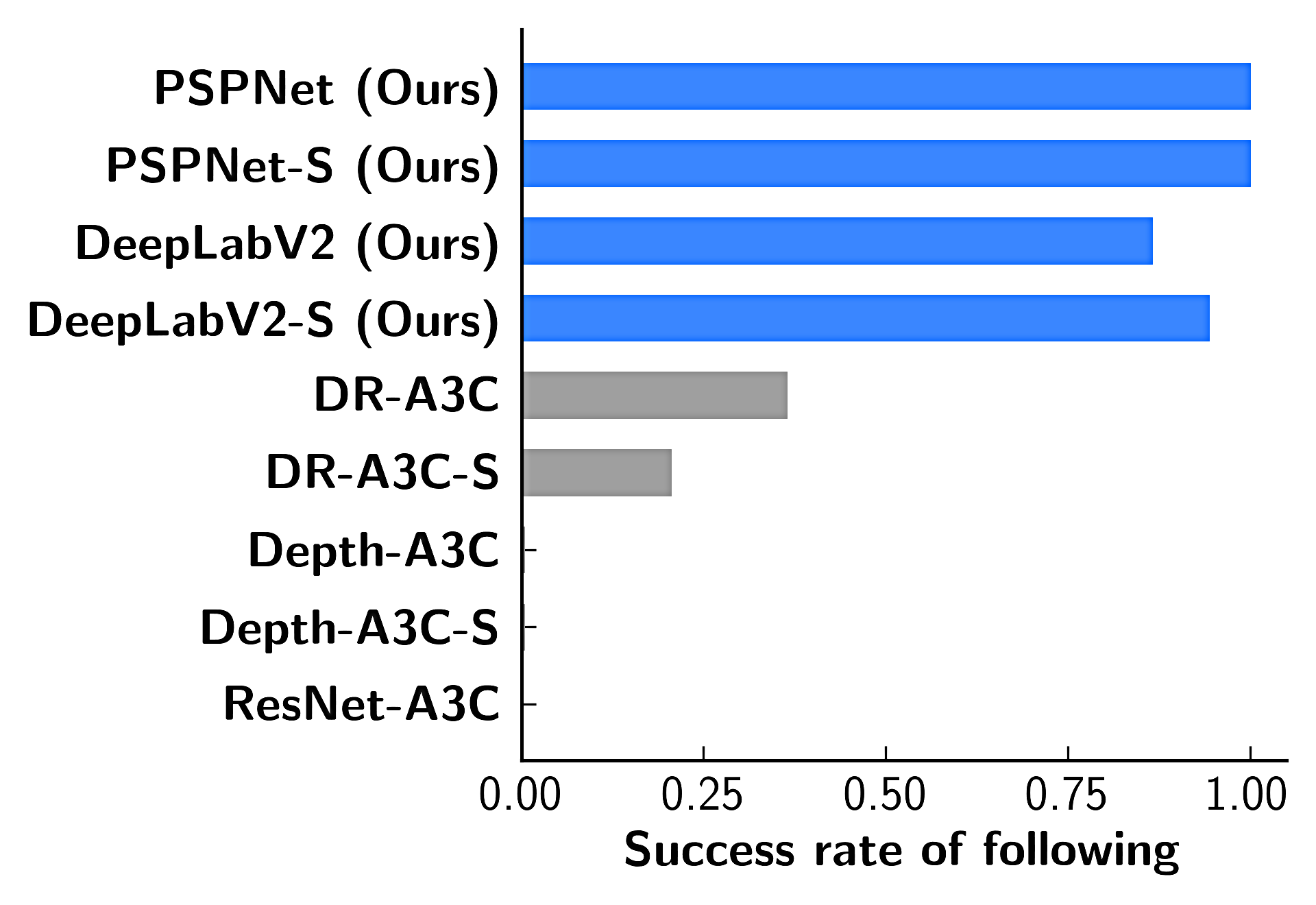}
\small
\captionof{figure}{Success rate in the target following task (evaluated in the real world).}
\label{figure::following_rate_real}
\end{figure}

\paragraph{Evaluation in the Real World}
To evaluate if a learned policy can be successfully transfered to the real world, we further conduct experiments in real indoor environments, which consist of corridors, forks, sharp turns, and randomly placed static obstacles. The target is a human who moves in the environment. The task is considered failed if the agent loses the target or collides with an obstacle, and successful if the agent follows the target to the destination. We report the success rate for each model in Fig.~\ref{figure::following_rate_real}. The success rate is evaluated over ten episodes. The results show that all the baseline models fail in the real-world scenarios. Specifically, as the models trained with DR never learn a useful policy even in the simulated training environments, there is no doubt that they fail in the real-world evaluation. As for the cases of Depth-A3C and Depth-A3C-S, despite the high mean rewards they attained in the training phase, the learned policies do not transfer successfully to the real world. During the evaluation, we observe that DR-A3C and ResNet-A3C agents treat the target person as an obstacle, and avoid the target as it comes close.
We also notice that our models do not lose track of the target even at sharp turns. We deduce that this is due to the temporal correlations preserved by frame stacking. 
Fig.~\ref{figure::following_rate_real} also provides evidence that our architecture is robust to the quality of image segmentation even in the target following task. 

\subsection{Demonstration of Visual Guidance}
\label{subsec::visualguide}

\begin{figure}[t]
\centering
\begin{subfigure}{.49\linewidth}
	\centering
	\includegraphics[width=0.7\linewidth]{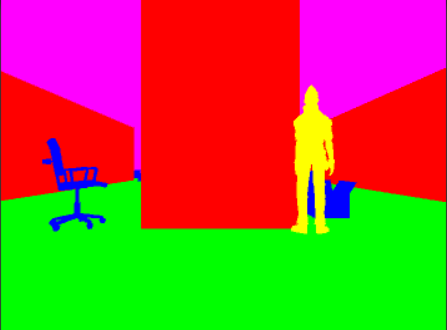}
	\caption{The target is a person}
	\label{fig:sw_person}
\end{subfigure}
\hfill
\begin{subfigure}{.49\linewidth}
	\centering
	\includegraphics[width=0.7\linewidth]{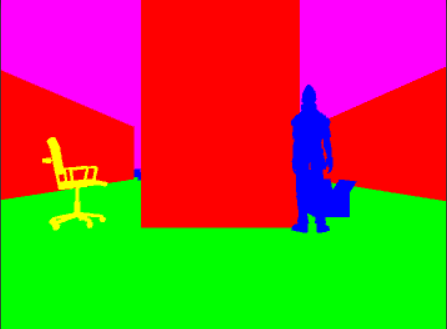}
	\caption{The target is a chair}
	\label{fig:sw_chair}
\end{subfigure}
\caption{Visual guidance allows the target to be switched.
}
\label{fig:sw_target}
\end{figure}

\begin{table}[t]
\renewcommand{\arraystretch}{1.25}
\centering
\scriptsize
\begin{tabulary}{1\linewidth}{l|C|C} \toprule
& VW (mean reward)	& RW (success rate) \\ \hline
Seg		& 0.824				& 80\% \\ \hline
Seg-S	& 0.925				& 90\% \\ \bottomrule
\end{tabulary}
\caption{Performance of visual guidance in the virtual world (VW) and real world (RW) in the \textit{Switching-Target Following} task.}
\label{table:exp_flexibility}
\end{table}

\begin{table}[t]
\centering
\begin{minipage}[t]{0.9\linewidth}
	\renewcommand{\arraystretch}{1.25}
	\centering
	\scriptsize
	\begin{tabulary}{1\linewidth}{L|C} \toprule
	Model						& Collision rate \\ \hline
	Reduced-Seg					& 0.7 \\ \hline
	Reduced-Seg-S				& 0.9 \\ \hline
	Non-Reduced-Seg				& 0.1 \\ \hline
	Non-Reduced-Seg-S			& 0.2 \\ \bottomrule
	\end{tabulary}
	\caption{Comparison of collision rate in \textit{Cluttered Hallway}}
	\label{table::redundant_labels}
\end{minipage}
\end{table}

We demonstrate the effectiveness of the visual guidance module of our modular architecture by switching the target in a \textit{Switching-Target Following} task. We perform experiments in both virtual and real environments. Both environments are the same as those in the \textit{Target Following} task.
To demonstrate visual guidance, we randomly select an object in the environment, and re-label it as the target at the beginning of each episode.  The initial location of the agent is chosen such that the selected target falls in the field of view (FOV) of the agent.  The selected object is rendered to yellow, as shown in Fig.~\ref{fig:sw_target}.
Fig.~\ref{fig:sw_target} illustrates the procedure of switching the target from Fig.~\ref{fig:sw_target}~(a) a human to Fig.~\ref{fig:sw_target}~(b) a chair.  We measure the mean rewards over 1,000 episodes in the virtual environments, and the success rate over ten episodes in the real world. Table~\ref{table:exp_flexibility} summarizes the results.  They reveal that our agents can successfully catch the randomly specified targets in both environments, no matter what target size and shape are. 


\subsection{Robustness to Non-Reduced Labels}
\label{subsec::ablation}
We further show that our control policy module can retain its performance, even if its input contains non-reduced labels (i.e., the original class labels in the ADE20K dataset). We conduct obstacle avoidance experiments in \textit{Cluttered Hallway}, as it is the most complex scenario for our agents. When training in the simulated environments, we include non-reduced class labels in semantic segmentations.
There are totally 27 non-reduced labels used for validating the robustness of our method. This enforces the agents to learn more labels in the training phase. Given the same amount of training time, the agents are able to obtain similar learning curves as those shown in Fig.~\ref{figure::learning_curve_obstacle}. For evaluation in the real world, we compare the agents trained with non-reduced labels to those trained with reduced labels (i.e., the agents mentioned in Section \ref{subsec::obstacle_avoidance}). 
We adopt PSPNet as the perception module for both agents, and modify its output labels to be matched with the labels rendered by our simulator.  The experimental results in Table \ref{table::redundant_labels} show that the agents trained with non-reduced labels demonstrate lower collision rate than their counterparts mentioned in Section \ref{subsec::obstacle_avoidance}. We attribute this improvement to the richer context of 
image segmentations 
perceived by the agents in the training phase.

\section{Conclusion}
\label{sec::conclusion}
In this paper, we presented a new modular architecture for transferring policies learned in simulators to the real world for vision-based robotic control. We proposed to separate the model into a perception module and a control policy module, and introduced the concept of using semantic image segmentation as the meta state for relating these two modules. We trained our model with a standard RL algorithm, and did not apply any DR technique. We performed experiments in two benchmark tasks: an obstacle avoidance task and a target following task, and demonstrated that the proposed method outperforms the baseline models in both virtual and real environments. 
We further validated the generalizability of our model in handling unfamiliar indoor and outdoor scenarios, and the transferability of our model from simulation to the real world without fine-tuning. Finally, in the switching-target following task, we proved that our model is flexible such that the target can be easily switched by visual guidance.

\section*{Acknowledgments}
\label{Acknowledgments}
The authors thank Lite-On Technology Corporation for the support in researching funding, and NVIDIA Corporation for the donation of the Titan X Pascal GPU used for this research.

\bibliographystyle{named}
\bibliography{ijcai18}

\end{document}